\documentclass{article}

\usepackage{arxiv}

\usepackage[utf8]{inputenc} % allow utf-8 input
\usepackage[T1]{fontenc}    % use 8-bit T1 fonts
\usepackage{hyperref}       % hyperlinks
\usepackage{url}            % simple URL typesetting
\usepackage{booktabs}       % professional-quality tables
\usepackage{amsfonts}       % blackboard math symbols
\usepackage{nicefrac}       % compact symbols for 1/2, etc.
\usepackage{microtype}      % microtypography
\usepackage{lipsum}
\usepackage{verbatim}
\usepackage{array}
\usepackage{multirow}
\usepackage{graphicx}
\usepackage{amssymb,amsmath}
\usepackage{pgf}
\usepackage{tikz}
\usepackage{bbm}
\usepackage{hyperref}
\usepackage[numbers]{natbib}

\newcommand{\CA}{\texttt{EQUAL}}
\newcommand{\CB}{\texttt{HOMOGENEOUS}}
\newcommand{\CC}{\texttt{HETEROGENEOUS}}

\title{Aligning with Heterogeneous Preferences for \\Kidney Exchange}

\author{
  Rachel Freedman\\
  University of California, Berkeley\\
  \texttt{rachel.freedman@berkeley.edu}
}

\graphicspath{ {fig/} }

\begin{document}

\maketitle

\begin{abstract}

  AI algorithms increasingly make decisions that impact entire groups of humans. Since humans tend to hold varying and even conflicting preferences, AI algorithms responsible for making decisions on behalf of such groups encounter the problem of \textit{preference aggregation}: combining inconsistent and sometimes contradictory individual preferences into a representative aggregate. In this paper, we address this problem in a real-world public health context: \textit{kidney exchange}. The algorithms that allocate kidneys from living donors to patients needing transplants in kidney exchange matching markets should prioritize patients in a way that aligns with the values of the community they serve, but allocation preferences vary widely across individuals.
  In this paper, we propose, implement and evaluate a methodology for prioritizing patients based on such \textit{heterogeneous} moral preferences. Instead of selecting a single static set of patient weights, we learn a distribution over preference functions based on human subject responses to allocation dilemmas, then sample from this distribution to dynamically determine patient weights during matching. We find that this methodology increases the average rank of matched patients in the sampled preference ordering, indicating better satisfaction of group preferences. We hope that this work will suggest a roadmap for future automated moral decision making on behalf of heterogeneous groups.
\end{abstract}

\section{Introduction}

    As AI algorithms become increasingly powerful and more widely deployed, it is vital that they act in a way that aligns with human values. Unfortunately, in most real-world domains, people do not unanimously agree on a single set of ``human values'' that AI algorithms can model and instantiate. Instead, groups of humans tend to hold varying and even conflicting moral preferences, and AI algorithms responsible for making decisions on behalf of these \textit{heterogeneous} groups must aggregate and arbitrate between these preferences. 
    
    Many existing approaches to \textit{preference aggregation} for AI rely on determining a single representative objective or decision for the AI to implement~\cite{lee2018webuildai,zhang2019pac,conitzer2015crowdsourcing}. However, humans are known for their variable and contradictory preferences, meaning that many individuals will hold preferences that differ greatly from the mean. Better techniques are required to model such heterogeneous human preferences, implement them in AI algorithms, and measure their satisfaction in practice.

    One domain in which this is particularly apparent is that of \textit{kidney exchange}. In a kidney exchange, patients who need kidney transplants and have found willing but medically incompatible donors are matched and exchange kidneys with other such incompatible patient-donor pairs~\cite{roth2004kidney}. Many countries, including the United States~\cite{dickerson2015futurematch}, the United Kingdom~\cite{manlove15paired} and much of Europe~\cite{biro17kidney} use algorithms developed in the AI community to automate this matching. Since the prognosis for patients who do not receive kidney transplants is quite poor, these automated decisions have life-or-death consequences and great moral import. It is therefore vital that these allocation decisions are made in a way that aligns with societal values. Previous work has sought to learn a single static utility function that kidney allocation algorithms can use to prioritize certain types of matching~\cite{freedman2020adapting}. However, that work disregards the empirical heterogeneity in human ethical judgements in this domain. We seek to instead model this heterogeneity by developing a methodology that represents the full distribution of human judgements.

    In this work, we draw on preference aggregation and social welfare theory to design, implement and evaluate a methodology for autonomously allocating kidneys to patients in matching markets based on the heterogeneous moral preferences expressed by surveyed human participants. We propose an alternate model for aggregating preferences drawn from the economics literature and an alternate domain-specific measure of group welfare based on individual preference rankings. We show that our proposed model, which aggregates individual preferences into a distribution over utility functions instead of a single function, improves the average rank of the matched kidney donations in individuals' preference orderings, without reducing the number of patients that can be matched overall. Incorporating the model into this real-world AI system leads to more beneficial outcomes according to our proposed measure of social welfare. 
    
    We hope that this work will both highlight the preference aggregation challenges present in many allocative AI systems, and serve as a roadmap for developing systems that directly address these challenges in other real-world contexts.

    % What is the problem?
        % In kidney exchanges, patients with willing but incompatible donors are matched.
        % In many countries, this matching is done by AI algorithms
        % Because this is a life/death situation, the decisions have moral consequences
        % Should align with human values
        % Preliminary work in this area, but tries to condense heterogeneous preferences into a single set of "scores"
        % Preferences are actually heterogeneous
    
    % Why is studying it important?
        % Many AI systems involve allocations with moral implications
        % Peoples' preferences are heterogeneous
        % We need better ways to model heterogeneous preferences and measure of their satisfaction
    
    % What did you do?
        % Propose an alternate model for preferences (BLP) and measure of satisfaction (rank)
        % Simulate matching using this and baseline (single set of scores, and all equal) versions
        % Show that BLP produces significantly different matching with improved average rank

    % What is your key insight?
        % Propose better way to deal with heterogenous preferences
    
    % Why are your results important?
        % Improve kidney allocation in short term
        % Suggest techniques and considerations for automating moral decision-making by generalizing from heterogeneous human value judgements
        
\section{Kidney Exchanges}

    \subsection{Graph Formulation}
        In a \textit{kidney exchange}, patients who need a kidney transplant and donors who are willing to donate to them but are medically incompatible can be matched with other such patient-donor pairs~\cite{roth2004kidney}. For example, if the donor of pair $i$ is compatible with the patient of pair $j$, and the donor of pair $j$ is likewise compatible with the patient of pair $i$, they can form a \textit{matching}, in which donor $d_i$ donates to patient $p_j$ and donor $d_j$ donates to patient $p_i$. 
         
\begin{figure}
            \centering
            \includegraphics[width=.7\linewidth]{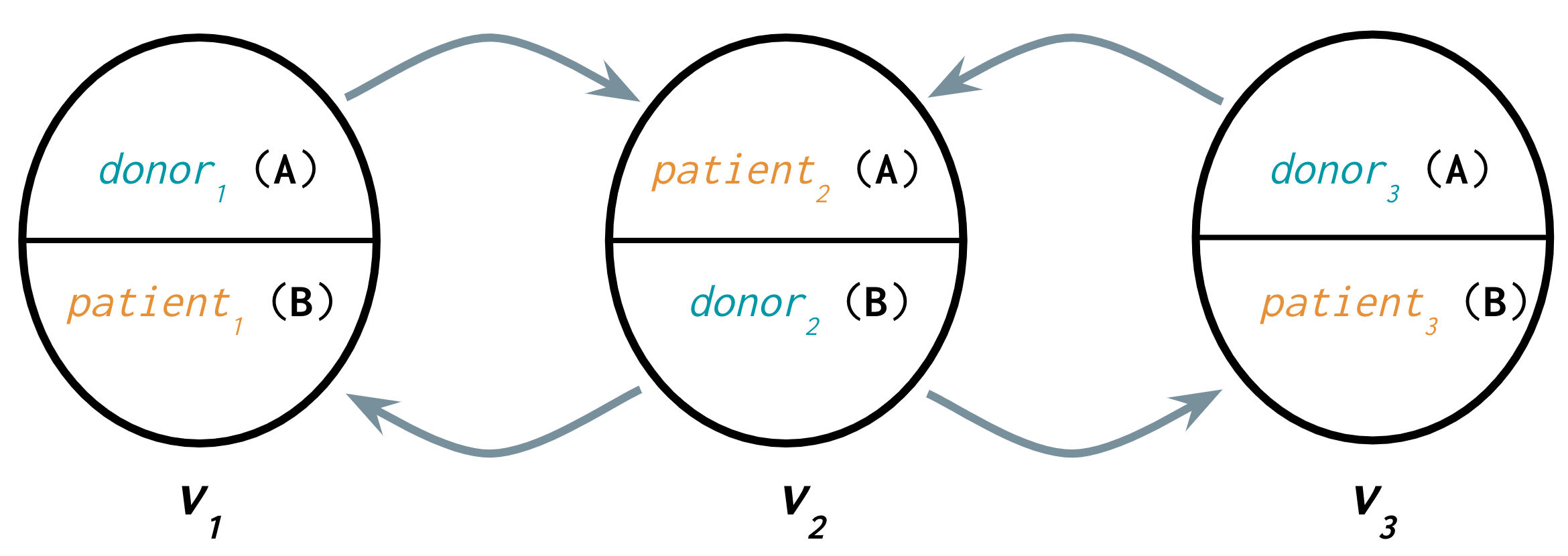}
            \caption{Example compatibility graph. Donor and patient blood types are in parentheses and arrows indicate possible valid donations. This graph has two valid donation cycles: $c_A = \{v_1, v_2\}$ and $c_B = \{v_2, v_3\}$. However, both contain $v_2$, so only one set of donations can take place. }
            \label{fig:KE}
        \end{figure}

        In the standard formulation, a kidney exchange is described by a \textit{compatibility graph} $G = \langle V, E\rangle$~\cite{roth2004kidney,roth2005kidney}. We construct one vertex $v$ for each patient-donor pair, then add a directed edge $e_{i,j}$ from $v_i$ to $v_j$ if $d_i$ is compatible with $p_j$. A cycle $c$ is a possible sequence of valid transplants, in which each donor in the cycle donates a kidney and each patient receives one. A matching $M$ is a set of disjoint cycles. An example compatibility graph is shown in Figure~\ref{fig:KE}. Each oval in the figure represents a vertex, and each arrow represents a directed edge signifying donor-patient compatibility. There are two cycles in this particular compatibility graph: $c_A = \{v_1, v_2\}$ and $c_B = \{v_2, v_3\}$. However, these two cycles are not disjoint, because they share vertex $v_2$. The $v_2$ donor cannot donate both of their kidneys, so these exchanges cannot both take place. This compatibility graph therefore has two valid matchings, $M_A = \{c_A\}$ and $M_B = \{c_B\}$, each with cardinality 2.
        
    \subsection{Clearing Algorithm}
        The \textit{clearing house problem} in kidney exchange is to find the optimal valid matching, according to some utility function~\cite{abraham2007clearing}. Finding valid matchings with a finite limit on cycle lengths is NP-hard~\cite{abraham2007clearing} and difficult to approximate~\cite{biro2007inapproximability}, so this problem is typically solved by formulating it as a linear program (LP) and solving it with an LP-solver such as CPLEX.
        
        We typically assign a weight $w_e$ to each edge $e$ to represent the utility of that particular donation taking place. In the national US exchange, these weights are set ad-hoc by a committee~\cite{UNOS15:Revising}, but in this work we will adapt an alternative method that learns these weights based on human responses to allocation dilemmas~\cite{freedman2020adapting}. The clearing house problem is to find the optimal matching $M^*$ which maximizes some utility function $u : M \rightarrow \mathbb{R}$. This is typically formalized as the graph-theoretic problem of finding the maximum weighted cycle cover $u(M) = \sum_{c \in M} \sum_{e \in c} w_{e}$. To solve this via linear programming, let $C(L)$ be the set of all cycles of length no more than $L$, let $w_c = \sum_{c\in C(L)}w_e$, create an activation variable $x_c \in {0, 1}$ for each cycle $c\in C(L)$, then solve the following linear program:
        
            \begin{equation}\label{eq:ilp}
                \max \sum_{c \in C(L)} w_c \ x_c \qquad
                \mathit{s.t.} \qquad
                \sum_{c : v \in c} x_c \leq 1 \quad \forall v \in V.
            \end{equation}

        \noindent using an LP-solver such as CPLEX. The final matching $M$ is the set of cycles $c$ with an activation $x_c = 1$. If all edge weights $w_e$ are equal, then solving this LP gives a maximum-cardinality matching. In cases where there are multiple valid matchings of equivalent cardinality, such as $M_A$ and $M_B$ in Figure~\ref{fig:KE}, this LP-solver must choose between them randomly. However, if the edge weights are set according to some utility function, then the solution can prioritize certain types of matches. This allows us to incorporate societal preferences.
    
    \subsection{Incorporating Preferences}\label{sec:prior}
    
        \begin{table}
            \begin{center}
                \begin{tabular}{ cccc } \\\toprule
                    Profile & Age & Drinking & Cancer \\ \midrule\midrule
                    1 & 30 & rare & healthy \\ \midrule
                    2 & 30 & frequently & healthy \\ \midrule
                    3 & 30 & rare & cancer \\ \midrule
                    4 & 30 & frequently & cancer\\ \midrule
                    5 & 70 & rare & healthy \\ \midrule
                    6 & 70 & frequently & healthy \\ \midrule
                    7 & 70 & rare & cancer \\ \midrule
                    8 & 70 & frequently & cancer \\ \bottomrule
                \end{tabular}
                \vspace{10pt}
                \caption{Patient profile descriptions enumerated and (arbitrarily) numbered by \citeauthor{freedman2020adapting}}
                \label{tab:profiles}
            \end{center}
        \end{table}
    
        Previous work has attempted to improve the matching prioritization in kidney exchanges based on sampled human ethical preferences~\cite{freedman2020adapting}. All else equal, it is obviously morally preferable to save lives by matching as many patient-donor pairs as possible. However, in cases such as the one in Figure~\ref{fig:KE}, there can be multiple maximum-cardinality matchings. In this case, the algorithm requires a utility function that distinguishes between them, ideally in a way that aligns with human values. The US national kidney exchange attempts to do this, but they prioritize matches in an opaque and ad-hoc fashion via committee~\cite{UNOS15:Revising}. This excludes most of the societal members who will actually participate in the exchange from the discussion, and leaves the committee with the still-unsolved problem of designing a utility function that captures the ethical preferences of an entire society. \citeauthor{freedman2020adapting} propose an alternative methodology for \textit{learning} domain-relevant ethical preferences from actual human decisions in kidney allocation dilemmas, revising the LP in Eq~\ref{eq:ilp} to take these into account, and then evaluating the impact on a simulated exchange. Our work proposes an improvement on \citeauthor{freedman2020adapting}'s methodology for aggregating preferences and evaluating results, so we will briefly outline their full methodology here.
        
        \citeauthor{freedman2020adapting} conducted two surveys on participants from Amazon's Mechanical Turk platform (``MTurk''). The first survey asked participants ($N = 100$) to read a brief description of the kidney transplant waiting list process, and then asked them to propose which patient characteristics they thought ``morally ought'' to be used to prioritize patients. The three most popular categories of responses were ``age'', ``health -- behavioral'' (including aspects of health believed to be under personal control, such as diet and drinking), and ``health -- general'' (including aspects of health unrelated to kidney disease). The second survey asked a new set of participants ($N = 289$) to decide how to allocate kidneys between pairs of fictional patient ``profiles'' that vary according to these attributes. In order to make the profiles more concrete, drinking behavior (``1 alcoholic drink per month'' or ``5 alcoholic drinks per day'') was used as a proxy for behavioral health, and cancer (``skin cancer in remission'' or ``no other major health problems'') was used as a proxy for general health. For example, a sample question asked participants to choose between ``Patient W.A. [who] is 30 years old, had 1 alcoholic drink per month (prior to diagnosis), and has no other major health problems'' and ``Patient R.F. [who] is 70 years old, had 5 alcoholic drinks per day (prior to diagnosis), and has skin cancer in remission''. They defined 8 such patient profiles, with characteristics described in Table~\ref{tab:profiles}, and asked each participant to compare each pair of profiles. We will use these profile descriptions and this preference data in our own work.
        
        \citeauthor{freedman2020adapting} used the \textit{Bradley-Terry Model} (``BT Model'') to estimate a ``BT-score'' for each patient profile. The BT model assumes that each profile $x$ has an underlying score $p_x$ that represents the value that survey participants collectively place on donating to a patient with that profile. Under this model, the probability that profile $i$ will be preferred to profile $j$ is:
            
            \begin{equation}
                P(i>j) = \frac{p_i}{p_i+p_j}
            \end{equation}
            
        \noindent Patient profiles that are almost always selected by our survey participants (such as profile 1 in Table~\ref{tab:profiles}) will therefore have the highest scores, and profiles that are rarely selected (such as profile 8), will therefore have the lowest scores. \citeauthor{freedman2020adapting} use this model to estimate a single set of scores based on the pooled pairwise comparisons from every survey respondent. This allows them to aggregate all preferences into a single set of scores.
        
        They then revised the LP from Eq~\ref{eq:ilp}, setting the weight of each edge $e_{i,j}$ to be the BT-score of the recipient $p_j$ and adding a cardinality constraint to require that the LP still only produce maximum-cardinality matchings. Let $w_{BT(v)}$ be the BT-score of the patient in vertex $v$, and let $Q$ be the maximum matching cardinality possible for the compatibility graph. Then the revised LP is:
        
            \begin{equation}
                \begin{array}{lll}
                    \max & \sum_{c \in C(L)} \left[ \sum_{(u,v) \in c} w_{BT(v)} \right] \ x_c & \\
                    \mathrm{s.t.} & \sum_{c : v \in c} x_c \leq 1 & \forall v \in V \\
                    & \sum_{c \in C(L)} |c| x_c \geq Q & \\
                \end{array}
                \label{eq:homogeneous}
            \end{equation}
        
        They evaluated this revised algorithm on a simulated kidney exchange and found that it matched significantly more of the higher-scoring profiles and significantly fewer of the lower-scoring ones. However, this methodology relies on the assumption that societal preferences are sufficiently homogeneous to be captured by a single static utility function. An algorithm using this methodology will always choose to save a patient of profile 1 over a patient of profile 2. However, the preferences expressed in the survey data actually varied greatly, and participants did sometimes prefer patients of profile 2 to profile 1. Presumably the preferences of a representative sample of the actual US population would be even more heterogeneous. In this sense, both the static profile scoring and the assessment of the algorithm by the proportion of each profile matched are flawed. In this work, we improve upon both of these elements by removing the requirement for a single utility function and developing an alternate methodology for modifying and evaluating the algorithm.
    
\section{Incorporating Heterogeneous Preferences}\label{sec:our}

    In our work we improve upon the methodology presented in Section~\ref{sec:prior} by removing the unrealistic assumption that societal preferences can be captured by a single utility function. We propose 1) an alternative preference aggregation method that better captures the variation in expressed preferences, 2) modifications to the kidney allocation algorithm to take this new preference aggregation into account, and  3) an alternative evaluation metric for the resulting matchings that lends more consideration to individual welfare. 

    \subsection{Preference Aggregation Model: BLP}\label{sec:blp}
    
        Instead of learning a single score for each profile as in previous work~\cite{freedman2020adapting}, we use the \textit{Berry-Levinsohn-Pakes Model} (``BLP Model'') to estimate a \textit{distribution} over possible utility functions. We propose that learning and sampling from this distribution better satisfies individual preferences than learning a single utility function. The BLP model is an extension of the logit discrete choice model that is widely used in estimating consumer discrete-choice demand for differentiated products~\cite{berry1995automobile,nevo2001measuring}. When we apply this model to kidney exchange, the ``consumers'' are members of the population that the exchange serves, and the ``products'' are the patients who may potentially be matched with donors. Using this model allows us to predict how the general population wants the exchange to prioritize patients.
        
        For a graph $G = \langle V, E\rangle$, we wish to define a utility function $\mathcal{U} : V \rightarrow \mathbb{R}$ that determines the utility of the patient in each vertex receiving a utility function. The BLP model defines the utility function $\mathcal{U}(v) = X_{p(v)}^\top \beta + \epsilon$ where $X_{p(v)} = \{\mathbbm{1}(v_{age} = 30), \mathbbm{1}(v_{drinking} = rare), \mathbbm{1}(v_{cancer} = healthy)\}$ are the binary features of the patient profile of vertex $v$, $\beta \sim \mathcal{N}(\mu, \Sigma)$ gives the weight of each feature, and  $\epsilon$ is an error term following a type-II extreme value distribution. 

        We use maximum likelihood estimation to fit the distribution parameters $(\mu, \Sigma)$ to the pairwise comparison survey data gathered by \citeauthor{freedman2020adapting} Let $P$ be the set of all patient profiles described in Table~\ref{tab:profiles}, and for each pair of profiles $i,j \in P$, let $c_k(i,j)$ be survey respondent $k$'s preferred profile. This allows us to define the likelihood function
        
        \begin{equation}
            L_k(\mu, \Sigma \mid c_k) = \mathbb{E}_{\beta \sim \mathcal{N}(\mu, \Sigma)}\left[\prod_{i,j}{\frac{\exp(X_{c_k(i, j)}^\top \beta)}{\exp(X_i^\top \beta)+\exp(X_j^\top \beta)}}\right]
            \label{eq:likelihood}
        \end{equation}
        
        \noindent and to estimate the maximum likelihood distribution parameters
        
        \begin{equation}
            \hat{\mu}, \hat{\Sigma} = \underset{\mu, \Sigma}{argmax}{\frac{1}{N}\sum_{k=1}^{N}\mathrm{log}({L_k(\mu, \Sigma \mid c_k))}}
            \label{eq:mle}
        \end{equation}
        
        % The estimates of the parameters are as follows:
        
        % \begin{equation*}
        %     \hat{\mu} = (8.18, 5.69, 3.53)^\top
        %     \label{eq:estimate_mu}
        % \end{equation*}
        
        % \begin{align*}
        %     &(\hat{\Sigma}_{11}, \hat{\Sigma}_{12}, \hat{\Sigma}_{13}) = (20.47, 2.54, 4.56), \\
        %     & (\hat{\Sigma}_{21}, \hat{\Sigma}_{22}, \hat{\Sigma}_{23}) = (2.54, 11.07, 1.30), \\
        %     & (\hat{\Sigma}_{31}, \hat{\Sigma}_{32}, \hat{\Sigma}_{33}) = (4.56, 1.30, 7.16)
        %     \label{eq:estimate_Sigma}
        % \end{align*}
        
    \subsection{Algorithm}
    
        % We use the same algorithm (2 LPs) as Freedman et al
        % Except generate each patient/donor pair with beta, and use this to assign weights and ranks for outward edges
        
        Each time a new patient-donor pair enters the exchange, we add a corresponding vertex $u$ to the graph, randomly sample a $\beta_{u} \sim \mathcal{N}(\hat{\mu}, \hat{\Sigma})$ from the learned distribution, and weight outgoing edges $u \rightarrow v$ using the resulting ``BLP function'': $BLP(u,v) = X_{p(v)}^\top \beta_u + \epsilon_{uv}$. In this way, we represent the full distribution of preferences. Note that the BLP function indicates a random sample from the surveyed population's preference distribution -- it does not represent the preferences of donor $u$ specifically. Letting $w_{BLP(u,v)}$ be the score that vertex $u$'s sampled BLP function places on donating to the patient in vertex $v$, we modify the LP in Eq~\ref{eq:homogeneous} to be:
        
            \begin{equation}
                \begin{array}{lll}
                    \max & \sum_{c \in C(L)} \left[ \sum_{(u,v) \in c} w_{BLP(u,v)} \right] \ x_c & \\
                    \mathrm{s.t.} & \sum_{c : v \in c} x_c \leq 1 & \forall v \in V \\
                    & \sum_{c \in C(L)} |c| x_c \geq Q & \\
                \end{array}
                \label{eq:heterogeneous}
            \end{equation}
          
    \subsection{Evaluation Metric}\label{sec:metric}
    
        We further define the $rank$ of a donation $u \rightarrow v$ to be the position of $v$'s patient profile in the preference ordering induced by $u$'s BLP function. For example, if the BLP function associated with vertex $u$ weights the profile of the patient in vertex $v$ above all other profiles, $rank(u,v) = 1$. Conversely, if the BLP function weights the profile below the other seven possible patient profiles, $rank(u,v) = 8$. In this context, \textit{rank} functions as a proxy for individual welfare because it represents the extent to which an individual's domain-relevant values were fulfilled. We claim that the average rank of matched donations is a better measure of the extent to which an algorithm values individual welfare than the proportion of each profile matched because the ranks of all matches depend on the full BLP distribution. In contrast, the proportion matched measure relies on the false assumption that everyone's preferences are better satisfied if patients with higher BT-scores are matched more often.
        
        We run both algorithms on a simulated kidney exchange, along with a third algorithm that weights all donations equally as a baseline. We evaluate the resulting matchings both on the proportion of each profile matched, and on our proposed average rank measure. We find that our proposed algorithm consistently outperforms both others on the rank measure, suggesting that it better represents the full distribution of societal preferences.

\section{Experiments}

    \subsection{Conditions}
    
        We tested three versions of the matching algorithm: the baseline one that weights all donations equally, the one with a single utility function described in Section~\ref{sec:prior}, and one with a distribution over utility functions proposed in Section~\ref{sec:blp}. 
    
        \paragraph{Condition 1: \CA}
        
            The \CA\, algorithm matches kidney exchange participants using the LP in Eq~\ref{eq:ilp}. That is, it weights all participants equally and chooses randomly amongst the highest-cardinality matchings. We use this condition as a baseline because it describes the case in which ethical preferences are not incorporated into the algorithm at all.
        
        \paragraph{Condition 2: \CB}
            
            The \CB\, algorithm matches participants using the LP in Eq~\ref{eq:homogeneous}. It assigns edge weights based on the BT-score of the recipient, relying on the assumption that individual preferences are sufficiently homogeneous to be captured by a single static utility function. This is the algorithm proposed by \citeauthor{freedman2020adapting}. See Table~\ref{tab:weights} for the weights used in the \CA$\,$ and \CB$\,$ conditions.
        
        \paragraph{Condition 3: \CC}
        
            The \CC\, algorithm matches participants using the LP in Eq~\ref{eq:heterogeneous}. It samples a BLP function when each vertex is added to the graph, normalizes the scores produced by that function to the range $[0,1]$, and uses that function and the profile of the recipient to weight each new outgoing edge. This allows for the possibility that heterogeneous individual preferences are better captured by a distribution than by a single utility function. This is the novel algorithm that we propose in this work. 
        
        \begin{table}[h]
            \centering
            \begin{tabular}{ccc} \toprule
                Profile ID & \CA  & \CB  \\ \midrule\midrule
                1 & 1.000 & 1.000 \\ \midrule
                2 & 1.000 & 0.103 \\ \midrule
                3 & 1.000 & 0.236 \\ \midrule
                4 & 1.000 & 0.036 \\ \midrule
                5 & 1.000 & 0.070 \\ \midrule
                6 & 1.000 & 0.012 \\ \midrule
                7 & 1.000 & 0.024 \\ \midrule
                8 & 1.000 & 0.003 \\ \bottomrule
            \end{tabular}
            \vspace{10pt}
            \caption{Patient profile weights for the \CA\, and \CB\, experimental conditions. The \CA\, algorithm values all profiles equally, so all have weight $1.0$. However, the \CB\, algorithm weights profiles according to their BT-scores. The \CC\, algorithm samples BLP functions throughout matching and so does not have a static weight for each profile.}
            \label{tab:weights}
        \end{table}
    
    \subsection{Measures}
    
        We evaluate each algorithm according to both the measure we propose in Section~\ref{sec:our} and the measure used by \citeauthor{freedman2020adapting}
    
        \paragraph{Average Rank}
        
            The \textit{average rank} of a matching is the average rank of each donation in the matching, where $rank(u,v)$ of a donation $u \rightarrow v$ is as defined in Section~\ref{sec:metric}. Recall that lower ranks indicate higher preference satisfaction and, since there are 8 profiles, all possible ranks fall in the range $[1.0,8.0]$. For each run of each algorithm, we recorded the average rank of every matching in the simulation, then averaged these to get the average rank value for that algorithm. 
        
        \paragraph{Proportion Matched}
        
            The \textit{proportion matched} of a profile is the proportion of patients of that type that entered the kidney exchange pool and were subsequently matched. A proportion matched of $100\%$ means that all patients of that type were matched, and a proportion matched of $0\%$ means that none were. For each run of each algorithm, we recorded the number of patients with each profile that entered the pool and the number of patients of each profile that were eventually matched, and used this to calculate proportion matched.

    \subsection{Experimental Setup}
    
        We built a simulator\footnote{All code for this paper can be found in the \texttt{Variation} package of \texttt{github.com/RachelFreedman/KidneyExchange}} to mimic daily matching using the \CA, \CB, or \CC\, algorithms based on previously developed tools~\cite{dickerson2015futurematch,freedman2020adapting}. Each simulated ``day'', some pairs enter the pool, some pairs exit the pool, and then the matching algorithm is run on the pairs that remain. The unmatched pairs remain in the pool to potentially be matched in the future. For each algorithm, we executed 50 runs of 5 years of simulated daily matching, and recorded the average matching rank and profile proportions matched for each run.

\section{Results}

    \paragraph{Average Rank}
    
\begin{figure}
            \centering
            \includegraphics[width=0.7\linewidth]{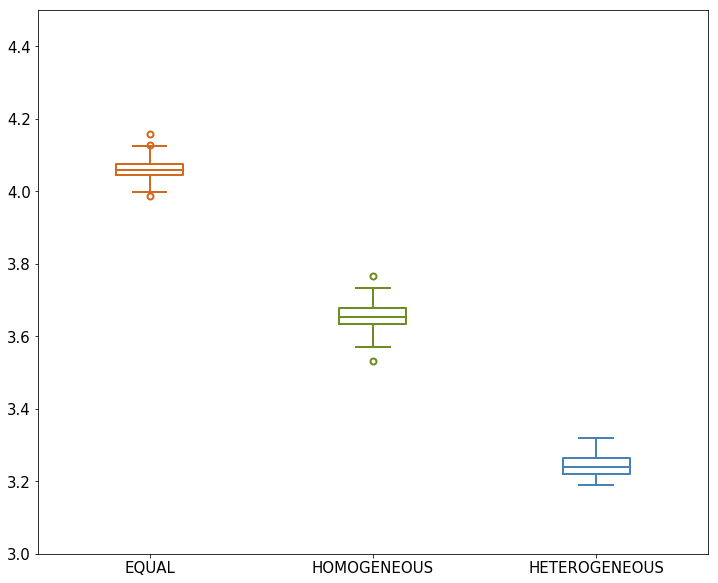}
            \caption{Average rank of donations in each simulated run ($N=50$). Lower ranks indicate more-preferred matches. The \CC\, algorithm produces the lowest average ranks (median = $3.24$), followed by the \CB\, algorithm (median = $3.66$), then the \CA\, algorithm (median = $4.06$).}
            \label{fig:average_rank}
        \end{figure}

        Since lower donation ranks indicate that the recipient is higher in the sampled preference ordering, we propose that algorithms that induce lower average ranks better satisfy population preferences. As expected, the proposed \CC\, algorithm consistently produces matchings with the lowest average rank (Figure~\ref{fig:average_rank}). The \CB\, algorithm produces the next-lowest average rank, followed by the \CA\, algorithm, which produces the highest average rank. This is because the \CA\, algorithm weights all edges \textit{equally}, matching recipients without any consideration of the personal characteristics used to define their weight. The \CB\, algorithm improves upon this by considering the characteristics of donation recipients, but fails to approximate preferences as closely as \CC.
        
    \paragraph{Proportion Matched}
    
\begin{figure}
            \centering
            \includegraphics[width=0.7\linewidth]{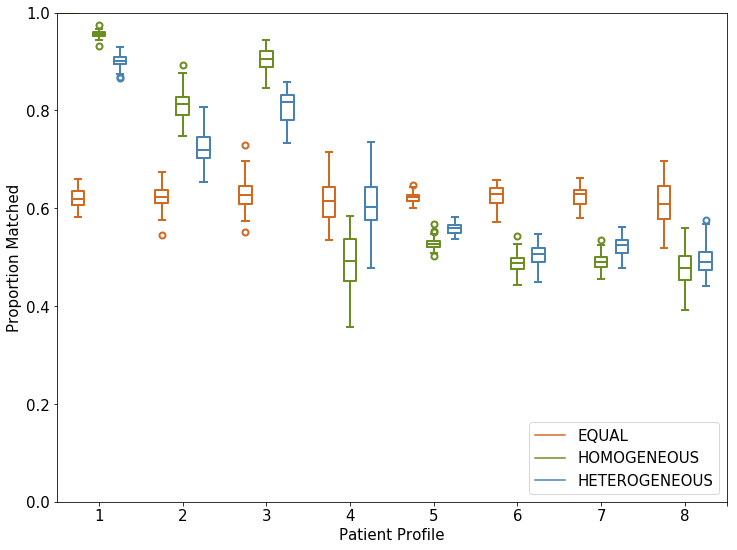}
            \caption{Proportion of patients of each profile matched in each simulated run ($N=50$). All algorithms match approximately $62\%$ of patients overall, but the \CB\, algorithm disproportionately matches profiles with higher BT-scores, the \CA\, algorithm matches all profiles approximately equally, and the proportions matched by the \CC\, algorithm lie between the other two.}
            \label{fig:proportion_matched}
        \end{figure}

        \citeauthor{freedman2020adapting} quantified the impact of their modified algorithm by comparing the proportions of patients of each profile type matched by their algorithm against the proportions matched by the unmodified algorithm, so for the sake of comparison we do the same. The proportions of each profile matched by the \CA\, algorithm, the \CB\, algorithm (proposed by \citeauthor{freedman2020adapting}) and the \CC\, algorithm (proposed in this work) are shown in Figure~\ref{fig:proportion_matched}. 
        
        Since it doesn't take patient profiles into account, the \CA\, algorithm matched approximately the same percentage of patients across all profiles. Since it prioritizes patients solely based on profile, the \CB\, algorithm matched the more popular profiles (1-3) more often and the less popular profiles (4-8) less. Notably, the \CB\, algorithm almost always matches patients of profile 1, indicating that a patient's profile can be one of the major factors in determining whether they receive a kidney. However, the \CC\, algorithm prioritizes patients not directly based on their profile, but based on the sampled BLP function's \textit{valuation} of their profile. As a result, this algorithm still tends to match more of the commonly-preferred profiles and fewer of the commonly-dispreferred ones, but sometimes samples a BLP function from the tails of the distribution that prioritizes patient profiles very differently. 
        
        If the survey preferences had been completely homogeneous, then the \CC\, algorithm would have produced the same results as the \CB\, algorithm. However, because preferences expressed in the survey data sometimes differ from the utility function used for the \CB\, algorithm, sometimes different matches are made. For example, while most survey participants preferred patient profile 1 to all other profiles, some did not, so the \CC\, algorithm respects this heterogeneity by sometimes prioritizing matching other profiles over profile 1. This difference in matching is a further indication that our proposed algorithm more faithfully represents the full distribution over preferences.
        
        % All three algorithms match the same proportion of patients overall, indicating that the modifications do not have any detrimental effect on matching.
        
\section{Discussion}

    % State how results show key insight.
    Faithfully instantiating the collective values of groups with heterogeneous individual preferences is a frequent challenge for real-world AI systems. For example, we commonly use AI systems to allocate scarce resources -- such as kidney donors, food donations~\cite{lee2018webuildai} and interview slots~\cite{schumann2019making} -- amongst group members in a way that we hope maximizes group welfare. Moreover, our roads may soon be populated with autonomous vehicles, which will have to make moral tradeoffs -- such determining who to sacrifice in unavoidable collisions~\cite{bonnefon2016social} -- based on the complex and often contradictory moral frameworks of the communities in which they operate. It is therefore vital that the AI community develop techniques for faithfully aggregating such heterogeneous preferences and use them to develop socially beneficial AI systems.
    
    In this paper, we proposed, instantiated, and evaluated one such technique for incorporating heterogeneous ethical preferences into a specific real-world AI system: an algorithm for matching patient-donor pairs in kidney exchange. Instead of weighting all potential kidney recipients equally, deciding how to prioritize them in an opaque and ad-hoc way~\cite{UNOS15:Revising}, or prioritizing them based on a single static utility function~\cite{freedman2020adapting}, we proposed learning a distribution over surveyed preferences and then sampling from this distribution for dynamic weighting during matching. We furthermore proposed donation rank as a better measure of preference satisfaction. We implemented our proposed algorithm and compare it to predecessor algorithms on a kidney exchange simulation, finding that our algorithm better satisfies survey participant preferences.
    
    Our model was estimated based on preference data elicited from MTurk survey participants, who are assuredly not representative of society in general. Future work should elicit preferences from a more representative sample, and perhaps privilege preferences expressed by domain experts and stakeholders such as doctors, policy-makers and kidney exchange participants. However, we believe that our sample was not \textit{more} heterogeneous than the US population as a whole, so we expect the challenge of heterogeneity and our methodology to continue to be relevant for this expanded sample. Moreover, since even a representative sample of the general public would still lack relevant domain-specific knowledge about kidney transplants, future work should also investigate methodologies that allow domain experts to correct or moderate the outcomes of this process.
    
    We hope that the challenges highlighted and methodology prototyped in this work will suggest a roadmap for developing techniques for automating moral decision making in other domains. 
    
\section*{Acknowledgements}

    We thank Yunhao Huang for implementing the BLP model, the Duke Moral AI team for sharing the human subject data, Dr. John Dickerson for building an earlier version of the kidney exchange simulation, and Dr. Peter Eckersley for early discussions of the idea.

\bibliographystyle{named}
\bibliography{refs}

\begin{thebibliography}{}

\bibitem[\protect\citeauthoryear{Abraham \bgroup \em et al.\egroup
  }{2007}]{abraham2007clearing}
David~J Abraham, Avrim Blum, and Tuomas Sandholm.
\newblock Clearing algorithms for barter exchange markets: Enabling nationwide
  kidney exchanges.
\newblock In {\em Proceedings of the 8th ACM conference on Electronic
  commerce}, pages 295--304, 2007.

\bibitem[\protect\citeauthoryear{Berry \bgroup \em et al.\egroup
  }{1995}]{berry1995automobile}
Steven Berry, James Levinsohn, and Ariel Pakes.
\newblock Automobile prices in market equilibrium.
\newblock {\em Econometrica: Journal of the Econometric Society}, pages
  841--890, 1995.

\bibitem[\protect\citeauthoryear{Biro and
  Cechlarova}{2007}]{biro2007inapproximability}
Peter Biro and Katarina Cechlarova.
\newblock Inapproximability of the kidney exchange problem.
\newblock {\em Information Processing Letters}, 101(5):199--202, 2007.

\bibitem[\protect\citeauthoryear{Bir{\'o} \bgroup \em et al.\egroup
  }{2017}]{biro17kidney}
P{\'e}ter Bir{\'o}, Lisa Burnapp, Bernadette Haase, Aline Hemke, Rachel
  Johnson, Joris van~de Klundert, and David Manlove.
\newblock Kidney exchange practices in {E}urope, 2017.
\newblock First Handbook of the {COST} {A}ction {CA}15210: {E}uropean {N}etwork
  for {C}ollaboration on {K}idney {E}xchange {P}rogrammes.

\bibitem[\protect\citeauthoryear{Bonnefon \bgroup \em et al.\egroup
  }{2016}]{bonnefon2016social}
Jean-Fran{\c{c}}ois Bonnefon, Azim Shariff, and Iyad Rahwan.
\newblock The social dilemma of autonomous vehicles.
\newblock {\em Science}, 352(6293):1573--1576, 2016.

\bibitem[\protect\citeauthoryear{Conitzer \bgroup \em et al.\egroup
  }{2015}]{conitzer2015crowdsourcing}
Vince Conitzer, Markus Brill, and Rupert Freeman.
\newblock Crowdsourcing societal tradeoffs.
\newblock In {\em Proceedings of the 2015 international conference on
  autonomous agents and multiagent systems}, pages 1213--1217, 2015.

\bibitem[\protect\citeauthoryear{Dickerson and
  Sandholm}{2015}]{dickerson2015futurematch}
John~P Dickerson and Tuomas Sandholm.
\newblock Futurematch: Combining human value judgments and machine learning to
  match in dynamic environments.
\newblock In {\em Twenty-Ninth AAAI Conference on Artificial Intelligence},
  2015.

\bibitem[\protect\citeauthoryear{Freedman \bgroup \em et al.\egroup
  }{2020}]{freedman2020adapting}
Rachel Freedman, Jana~Schaich Borg, Walter Sinnott-Armstrong, John~P Dickerson,
  and Vincent Conitzer.
\newblock Adapting a kidney exchange algorithm to align with human values.
\newblock {\em Artificial Intelligence}, page 103261, 2020.

\bibitem[\protect\citeauthoryear{Lee \bgroup \em et al.\egroup
  }{2018}]{lee2018webuildai}
Min~Kyung Lee, Daniel Kusbit, Anson Kahng, Ji~Tae Kim, Xinran Yuan, Allissa
  Chan, Ritesh Noothigattu, Daniel See, Siheon Lee, Christos-Alexandros Psomas,
  et~al.
\newblock Webuildai: Participatory framework for fair and efficient algorithmic
  governance.
\newblock {\em Preprint}, 2018.

\bibitem[\protect\citeauthoryear{Manlove and
  {O'M}alley}{2015}]{manlove15paired}
David Manlove and Gregg {O'M}alley.
\newblock Paired and altruistic kidney donation in the {UK}: Algorithms and
  experimentation.
\newblock {\em {ACM} Journal of Experimental Algorithmics}, 19(1), 2015.

\bibitem[\protect\citeauthoryear{Nevo}{2001}]{nevo2001measuring}
Aviv Nevo.
\newblock Measuring market power in the ready-to-eat cereal industry.
\newblock {\em Econometrica}, 69(2):307--342, 2001.

\bibitem[\protect\citeauthoryear{Roth \bgroup \em et al.\egroup
  }{2004}]{roth2004kidney}
Alvin~E Roth, Tayfun S{\"o}nmez, and M~Utku {\"U}nver.
\newblock Kidney exchange.
\newblock {\em The Quarterly journal of economics}, 119(2):457--488, 2004.

\bibitem[\protect\citeauthoryear{Roth \bgroup \em et al.\egroup
  }{2005}]{roth2005kidney}
Alvin~E Roth, Tayfun S{\"o}nmez, et~al.
\newblock A kidney exchange clearinghouse in new england.
\newblock {\em American Economic Review}, 95(2):376--380, 2005.

\bibitem[\protect\citeauthoryear{Schumann \bgroup \em et al.\egroup
  }{2019}]{schumann2019making}
Candice Schumann, Zhi Lang, Jeffrey Foster, and John Dickerson.
\newblock Making the cut: A bandit-based approach to tiered interviewing.
\newblock In {\em Advances in Neural Information Processing Systems}, pages
  4641--4651, 2019.

\bibitem[\protect\citeauthoryear{{UNOS}}{2015}]{UNOS15:Revising}
{UNOS}.
\newblock Revising kidney paired donation pilot program priority points, 2015.
\newblock OPTN/UNOS Public Comment Proposal.

\bibitem[\protect\citeauthoryear{Zhang and Conitzer}{2019}]{zhang2019pac}
Hanrui Zhang and Vincent Conitzer.
\newblock A pac framework for aggregating agents’ judgments.
\newblock In {\em Proceedings of the AAAI Conference on Artificial
  Intelligence}, volume~33, pages 2237--2244, 2019.

\end{thebibliography}

\end{document}